\documentclass[sigconf]{acmart}

\usepackage{amssymb}
\usepackage{amsmath,amsfonts,bm}

\def\eqref#1{equation~\ref{#1}}
\def\1{\bm{1}}

\DeclareMathAlphabet{\mathsfit}{\encodingdefault}{\sfdefault}{m}{sl}
\SetMathAlphabet{\mathsfit}{bold}{\encodingdefault}{\sfdefault}{bx}{n}

\usepackage{pifont}% http://ctan.org/pkg/pifont

\usepackage{amsmath}

\usepackage{amsfonts}
\usepackage{graphicx}
\usepackage{textcomp}
\usepackage{enumitem}
\usepackage{xcolor}
\usepackage{multirow}
\usepackage{soul}
\usepackage{tikz}
\usepackage{booktabs}
\usepackage{algorithm}
\usepackage{algpseudocode}
\usepackage{colortbl}
\usepackage{bm}
\usepackage{float}
\usepackage{filecontents}
\usepackage{CJK}

\usepackage{pifont}% http://ctan.org/pkg/pifont
\setcopyright{none}
\renewcommand\footnotetextcopyrightpermission[1]{} % removes footnote with conference information in first column

\begin{document}

\pagestyle{plain}
\title{Think Before You Prune: Self-Reflective Structured Pruning for Reasoning Language Models\\
}

\author{Ziyan Wang}
\affiliation{
  \institution{University of North Carolina at Charlotte}
  \city{Charlotte}
  \country{NC}
}
\email{zwang53@charlotte.edu}

\author{Enmao Diao}
\affiliation{
  \institution{DreamSoul}
  \city{Chengdu}
  \country{China}
}
\email{diao_em@hotmail.com}

\author{Qi Le}
\affiliation{
  \institution{University of Minnesota}
  \city{Minneapolis}
  \country{MN}
}
\email{le000288@umn.edu}

\author{Pu Wang}
\affiliation{
  \institution{University of North Carolina at Charlotte}
  \city{Charlotte}
  \country{NC}
}
\email{pu.wang@charlotte.edu}

\author{Guanchu Wang}
\affiliation{
  \institution{University of North Carolina at Charlotte}
  \city{Charlotte}
  \country{NC}
}
\email{gwang16@charlotte.edu}

\author{Minwoo Lee}
\affiliation{
  \institution{University of North Carolina at Charlotte}
  \city{Charlotte}
  \country{NC}
}
\email{minwoo.lee@charlotte.edu}

\author{Shu-ping Yeh}
\affiliation{
  \institution{Intel Corporation}
  \city{Santa Clara}
  \country{CA}
}
\email{shu-ping.yeh@intel.com}

\author{Li Yang}
\affiliation{
  \institution{University of North Carolina at Charlotte}
  \city{Charlotte}
  \country{NC}
}
\email{lyang50@charlotte.edu}

\begin{abstract}
Reasoning LLMs (RLMs) such as OpenAI o1, DeepSeek-R1, and Qwen3 deliver strong multi-step reasoning through chain-of-thought generation, but their large model sizes and lengthy decode-time outputs make them costly to deploy and unsuitable for resource-constrained settings. To reduce computing and memory cost, pruning offers a promising solution by removing unimportant parameters. However, despite their success on standard LLMs, existing pruning methods severely damage RLMs, as even moderate sparsity (e.g., 20\%) can collapse accuracy and completely disrupt the model’s reasoning coherence. We begin by analyzing why existing pruning pipelines fail on reasoning LLMs and find that their brittleness largely stems from a mismatch between the calibration data, the pruning objective, and the model’s decode-time reasoning behavior. Our study further shows that the most reliable calibration signal comes not from human-written labels but from the model’s own self-generated reasoning traces, which more accurately reflect its inference distribution. Guided by these insights, we introduce RESP, a self-reflective structured pruning framework that aligns pruning decisions with the model’s reasoning dynamics through self-generated calibration, decode-only gradient-based importance estimation, and progressive regeneration that maintains calibration fidelity as sparsity increases. Experiments on Qwen3-8B demonstrate that RESP markedly outperforms existing structured pruning methods on both GSM8K and MathQA, preserving near-dense accuracy at 20–30\% sparsity and substantially mitigating performance collapse at higher sparsity levels. At 40\% sparsity, RESP attains 81.3\% accuracy on GSM8K and 59.6\% on MathQA, surpassing the strongest baselines by 66.87\% and 47\%, respectively.
\end{abstract}

\received{18 November 2025}
%\received[revised]{12 March 2009}
%\received[accepted]{5 June 2009}

%%
%% This command processes the author and affiliation and title
%% information and builds the first part of the formatted document.
\maketitle

% \begin{IEEEkeywords}
% Structured Pruning; Reasoning Large Language Models; Calibration;
% \end{IEEEkeywords}

\section{Introduction}
Large language models (LLMs)~\cite{touvron2023llama,touvron2023llama2openfoundation,openai2024gpt4technicalreport,grattafiori2024llama3herdmodels} have achieved substantial progress in language understanding and generation. % yet next-token training on generic corpora favors fluency over deliberate multi-step reasoning. 
Building on these advances, 
Reasoning LLMs (RLMs), such as OpenAI o1~\cite{openai2024openaio1card}, DeepSeek-R1~\cite{deepseekai2025deepseekr1incentivizingreasoningcapability}, and Qwen3~\cite{yang2025qwen3technicalreport}, 
further strengthen reasoning ability and deliver significant progress on complex tasks, including reliable symbolic manipulation~\cite{wei2023chainofthoughtpromptingelicitsreasoning}, mathematical derivation~\cite{cobbe2021trainingverifierssolvemath,amini2019mathqainterpretablemathword}, or multi-hop inference~\cite{phan2025humanitysexam,suzgun2022challengingbigbenchtaskschainofthought}.
These models acquire their reasoning capability by explicitly producing chain-of-thought (CoT)~\cite{wei2023chainofthoughtpromptingelicitsreasoning} traces learned through supervised rationales~\cite{zelikman2022starbootstrappingreasoningreasoning} or reinforcement learning (RL)~\cite{schulman2017proximalpolicyoptimizationalgorithms,shao2024deepseekmathpushinglimitsmathematical}. However, the substantially large model size hinder deployment, and their lengthy multi-step chain-of-thought outputs incur high inference costs~\cite{zhang2025reasoningmeetscompressionunderstanding,sui2025stopoverthinkingsurveyefficient,xu2025largereasoningmodelssurvey,feng2025efficientreasoningmodelssurvey}, making these models expensive for individual users and impractical for resource-constrained devices.   

\begin{figure*}[t]
\centering
\resizebox{1\linewidth}{!}{
\includegraphics{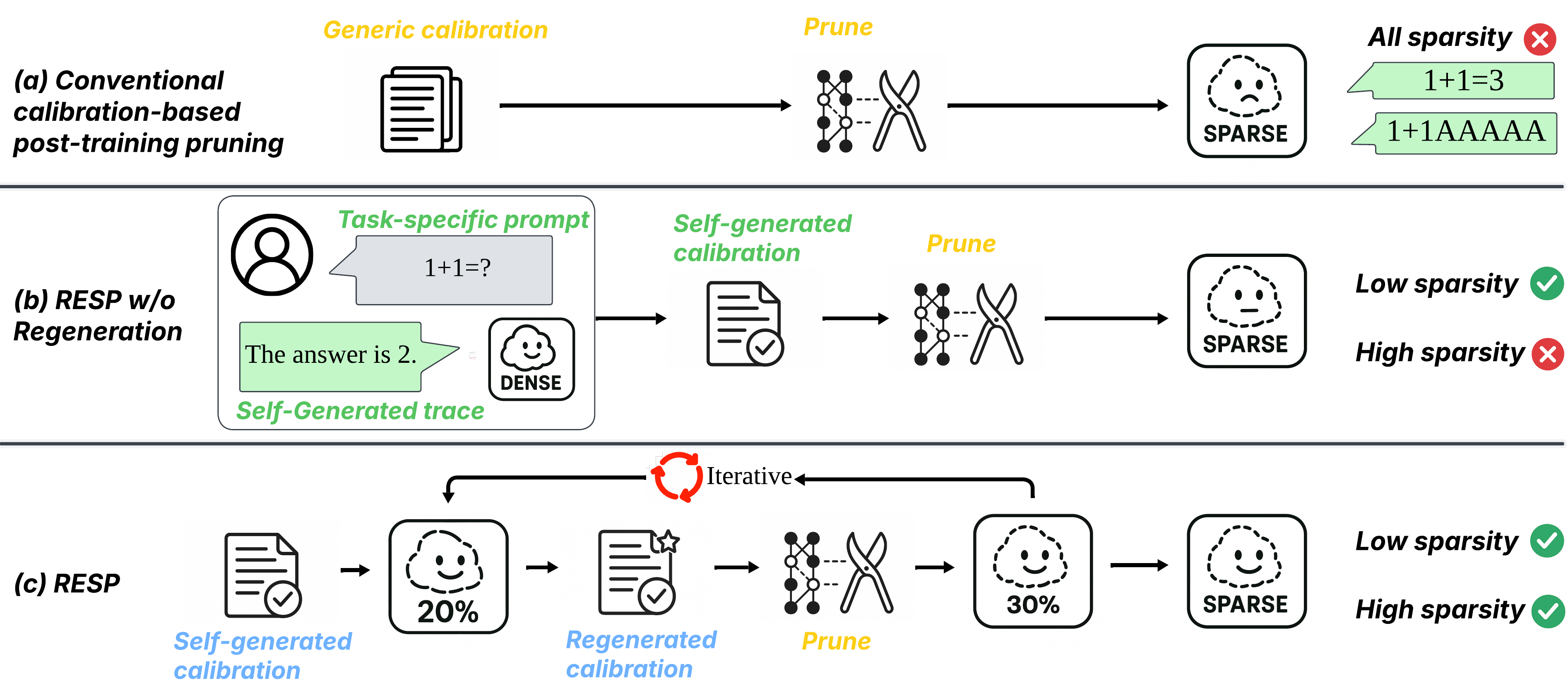}
}
\caption{Overview for RESP. \textbf{(a)} Conventional post‑training methods use generic corpora for calibration, causing severe reasoning degradation.
\textbf{(b)} RESP builds calibration from task‑specific prompts and self‑generated traces to better align the RLMs' decoding nature, achieving performance gain.
\textbf{(c)} With regeneration, RESP further boosts performance under high sparsity.}
\label{fig:overview}
\end{figure*}

Given these constraints, pruning has become a popular model compression method to reduce the computing and memory cost of LLMs by removing less important parameters. In practice, due to the substantial training cost, pruning for LLMs is predominantly performed in the post-training setting, where a lightweight calibration data~\cite {ma2023llm, frantar2023sparsegptmassivelanguagemodels, sun2024simpleeffectivepruningapproach, kim2024shortenedllamadepthpruning,an2023fluctuationbasedadaptivestructuredpruning,wang2025localglobalrevisitingstructured} is used to estimate the importance score for selecting which parameter to be pruned, as shown in Fig.\ref{fig:overview}(a). Upon that, existing post-training pruning methods for LLMs can be broadly categorized into \textit{OBS-based local methods} and \textit{gradient-based global methods}. OBS-based methods, beginning with Optimal Brain Surgeon (OBS) and its layer-wise variants~\cite{NIPS1992_303ed4c6,dong2017learningprunedeepneural}, decompose compression into per-layer subproblems and optimize a local $L_2$ reconstruction loss between the outputs of the unpruned and pruned layers. Recent works, such as SparseGPT~\cite{frantar2023sparsegptmassivelanguagemodels}, Wanda~\cite{sun2024simpleeffectivepruningapproach}, FLAP~\cite{an2023fluctuationbasedadaptivestructuredpruning}, and OWL~\cite{yin2025outlierweighedlayerwisesparsity}, instantiate this paradigm with different approximations of the layer-wise reconstruction objective. In contrast, gradient-based methods, including LLM-Pruner~\cite{ma2023llm}, GBLM~\cite{das2024sizegradientsshapepruning}, and GISP~\cite{wang2025localglobalrevisitingstructured}, measure parameters' importance globally by aggregating first-order Taylor saliency with respect to the model's final loss. In addition, based on the sparsity pattern, pruning methods can be divided into unstructured pruning and structured pruning. Unstructured pruning removes individual weights and offers high flexibility but provides limited real-world speedup, while structured pruning eliminates entire computational units (e.g., heads, channels, layers) and yields practical compute and memory reduction on general-purpose GPUs without relying on specialized accelerator designs. Given its hardware compatibility, in this work, we focus on structured pruning for reasoning LLMs.

Despite their effectiveness on general non-reasoning LLMs, existing pruning methods exhibit significant accuracy degradation when applied to RLMs. For example, at 30\% sparsity, existing calibration-based post-training pruning methods incur only 9.4\% accuracy degradation~\cite{wang2025localglobalrevisitingstructured} on the non-reasoning CommonsenseQA benchmark \cite{hu2023llmadaptersadapterfamilyparameterefficient}. In sharp contrast, the reasoning task GSM8K~\cite{cobbe2021trainingverifierssolvemath}, prior pruning methods cause accuracy to collapse to below 10\%, despite the dense Qwen3-8B model achieving 96.74\%. The pruned models make elementary arithmetic mistakes, generate repetitive or non-progressive chains of thought, and even produce confidently incorrect conclusions. This gap between existing pruning methods and RLMs motivates us to address the central question: \textbf{\textit{how to perform structured pruning in a way that preserves the reasoning capability of RLMs?}}

% on common multi-step math reasoning benchmarks GSM8K~\cite{cobbe2021trainingverifierssolvemath}, prior pruning methods the accuracy drop to less than 10\% at 30\% sparsity while the accuracy of the dense model is 96.74\% for Qwen3-8B, where the pruned models make elementary arithmetic mistakes, generate repetitive or non-progressive chains of thought, and even produce confidently incorrect conclusions. These observations motivate us to address the central question: \textbf{\textit{how to perform structured pruning in a way that preserves the reasoning capability of RLMs?}}

To understand why existing pruning methods fail on RLMs, we conduct a systematic study of calibration strategies and pruning algorithms, and uncover three key observations:
\textit{(1) Task-specific calibration is essential for preserving reasoning ability:} Generic corpora such as C4~\cite{raffel2023exploringlimitstransferlearning} could severely degrade accuracy, whereas calibration with reasoning-aligned data (e.g., GSM8K \cite{cobbe2021trainingverifierssolvemath}) substantially preserves performance at low sparsity.
\textit{(2) Gradient-based pruning benefits far more from task-specific calibration than OBS-based methods:} While OBS methods gain little beyond low sparsity, gradient-based GISP continues to recover substantial accuracy because its importance scores depend directly on task-specific gradients.
\textit{(3) Self-generated reasoning traces provide the most effective calibration signal:} Since RLMs are trained via reinforcement learning on their own generated sequences, their inference distribution is self-induced. Therefore, calibration using self-generated traces, rather than the human-written labels in the task-specific dataset, best matches this distribution, greatly improving gradient-based pruning.

Guided by these observations, we propose a \textbf{self-reflective structured pruning method for reasoning LLMs (RESP)}. As shown in Fig.\ref{fig:overview}(b), the core idea of RESP is to align pruning decisions with the model’s own reasoning behavior. To achieve this, RESP consists of three components. \textit{(1) Self-generated calibration:} Given prompts from a task-specific dataset, RESP constructs calibration data using the model’s own self-generated reasoning traces, ensuring that importance estimation is grounded in the distribution the model actually follows during multi-step reasoning. \textit{(2) Gradient-based importance estimation:} RESP adopts a gradient-based pruning strategy to compute importance scores for structured weights (e.g., attention heads or MLP channels). Critically, because reasoning accuracy is determined during the decode-time token generation stage rather than the prefill stage, RESP computes a decode-only loss that masks out all prompt tokens, ensuring that importance scores reflect decode-time reasoning behavior. \textit{(3) Progressive calibration regeneration:} We further observe that self-generated calibration collected from the dense model becomes mismatched at high sparsity (e.g., 40\%) due to a shift in the model’s reasoning distribution. To address this, RESP periodically regenerates fresh self-generated reasoning traces from the current pruned model as sparsity increases, keeping calibration aligned with the evolving model and preventing drift that would otherwise accumulate across pruning iterations (Fig.\ref{fig:overview}(c)).

In summary, this paper makes three contributions:
% \vspace{-2em}
\begin{itemize}
\item We empirically identify and characterize the reasons existing pruning pipelines fail on RLMs, revealing critical gaps in calibration data strategies, post-training pruning methods, and dependence on decode-time reasoning behavior.  

\item We propose RESP, a self-reflective structured pruning framework that leverages self-generated calibration, decode-time gradient importance, and progressive regeneration to preserve reasoning accuracy. To the best of our knowledge, this is the first work to address structured pruning for reasoning LLMs.

\item Extensive experiments on the latest open-sourced RLM Qwen3-8B across GSM8K and MathQA, demonstrating that RESP exceeds strong baselines at various sparsities and substantially extends the useful sparsity regime for RLMs, while remaining a fully post-training method. For example, RESP, for the first time, achieves near accuracy to dense models at 20–30\% sparsity and markedly delays collapse at higher sparsity. At 40\% sparsity, it achieves 81.3\% and 59.6\% accuracy on GSM8K and MathQA, surpassing the previous best baselines by 66.87\% and 47\%, respectively.
\end{itemize}

\section{Related Works}\label{sec:related_works}
\noindent\textbf{Reasoning Large Language Models.}
RLMs are LLMs optimized to generate intermediate chain-of-thought (CoT)~\cite{wei2023chainofthoughtpromptingelicitsreasoning} traces, which improve multi-step reasoning. They are typically obtained via supervised fine-tuning~\cite{zelikman2022starbootstrappingreasoningreasoning} on rationales or reinforcement learning on outcome signals (e.g., PPO~\cite{schulman2017proximalpolicyoptimizationalgorithms} and GRPO~\cite{shao2024deepseekmathpushinglimitsmathematical}). Contemporary systems such as o1~\cite{openai2024openaio1card}, DeepSeek-R1~\cite{deepseekai2025deepseekr1incentivizingreasoningcapability}, and Qwen3~\cite{yang2025qwen3technicalreport} exemplify this trend. However, the additional supervision and longer decode-time sampling raise compute and latency, spurring research on efficient reasoning and compression~\cite{snell2024scalingllmtesttimecompute,sui2025stopoverthinkingsurveyefficient,feng2025efficientreasoningmodelssurvey}.

\noindent\textbf{Pruning and challenge for RLMs.} 
Pruning removes redundant parameters by inducing sparsity. Classical work (e.g., OBD~\cite{NIPS1989_6c9882bb}) introduced Taylor-series–based importance estimation, later scaled across CNNs with strong empirical success~\cite{han2016deepcompressioncompressingdeep,molchanov2017pruningconvolutionalneuralnetworks,wang2021convolutionalneuralnetworkpruning}. With the advent of LLMs, pruning has re-emerged as a practical approach to efficient inference when full retraining is infeasible, motivating post-training methods that rely on lightweight calibration data rather than expensive end-to-end finetuning. 

In terms of sparsity pattern, pruning can be divided into \textit{unstructured} or \textit{structured}. Unstructured methods operate at the individual-weight level and keep the architecture intact. For example, SparseGPT  \cite{frantar2023sparsegptmassivelanguagemodels} casts pruning as a sparse regression problem and uses a Hessian-based approximate solver, while Wanda \cite{sun2024simpleeffectivepruningapproach} proposes a simpler importance score based on the product of weight magnitudes and input activations, achieving competitive quality with much lower cost. Structured pruning, in contrast, removes higher-level units such as attention heads, MLP channels, or entire layers, yielding compact architectures that are hardware-friendly and can realize wall-clock speedups without specialized kernels~\cite{wan2024efficientlargelanguagemodels,wang2024modelcompressionefficientinference,ma2023llm}. Representative methods include LLM-Pruner~\cite{ma2023llm}, which prunes heads and channels using gradient-based importance while respecting structural dependencies, as well as depth-pruning approaches such as ShortGPT~\cite{men2024shortgptlayerslargelanguage}, which remove layers with low layer-wise similarity scores.

% \begin{table}[tbp]
% \centering
% \caption{Calibration source matters for pruning reasoning LLMs. Calibrating on a \textbf{generic corpus (C4)} causes severe degradation, while \textbf{task-specific (GSM8K)} calibration preserves accuracy yet still collapses at high pruning ratios.}
% \label{tab:motivation_calib}
% \resizebox{\columnwidth}{!}{%
% \begin{tabular}{l|l|l|r|r}
% \toprule
% \textbf{Calibration Data} & \textbf{Category} & \textbf{Method} & \textbf{Pruning Ratio} & \textbf{GSM8K} \\
% \hline
% None & - & Dense & 0\% & 96.74\% \\
% \hline
% \multirow{9}{*}{C4} 
%  & \multirow{6}{*}{OBS} 
%  & \multirow{2}{*}{FLAP} & 20\% & 20.17\%
%  \\
%  &  &  & 30\% & 7.58\% \\
% \cline{3-5}
%  &  & \multirow{2}{*}{OWL} & 20\% & 23.96\% \\
%  &  & & 30\% & 1.82\% \\
% \cline{3-5}
%  &  & \multirow{2}{*}{Wanda} & 20\% &  25.40\%\\
%  &  & & 30\% & 2.81\% \\
% \cline{2-5}
%  & \multirow{3}{*}{Gradient} 
%  & \multirow{3}{*}{GISP} & 20\% &  47.31\%\\
%  &  & & 30\% & 9.25\% \\
%  &  & & 40\% & 6.52\% \\
% \hline
% \multirow{9}{*}{GSM8K} 
%  & \multirow{6}{*}{OBS} 
%  & \multirow{2}{*}{FLAP} & 20\% & 79.30\%
%  \\
%  &  & & 30\% &  49.58\% \\
% \cline{3-5}
%  &  & \multirow{2}{*}{OWL} & 20\% & 83.55\%
%  \\
%  &  & & 30\% & 8.64\% \\
% \cline{3-5}
%  &  & \multirow{2}{*}{Wanda} & 20\% & 85.90\% \\
%  &  & & 30\% & 16.38\% \\
% \cline{2-5}
%  & \multirow{3}{*}{Gradient} 
%  & \multirow{3}{*}{GISP} & 20\% & 90.52\% \\
%  &  & & 30\% & 79.68\%\\
%  &  & & 40\% & 14.40\%\\
% \bottomrule
% \end{tabular}%
% }
% \end{table}

\textit{Importance estimation.} Post-training pruning methods also differ in how they define the pruning objective and importance estimation.

\emph{(i) Local OBS-based (activation reconstruction).}
Following OBS \cite{NIPS1992_303ed4c6} and layer-wise OBS~\cite{dong2017learningprunedeepneural}, local methods decompose model compression into per-layer subproblems and minimize a reconstruction loss between the original and pruned layer outputs:
\begin{equation}
\small
\min_{M_{\ell},\,\widehat{W}_{\ell}}
\bigl\|
W_{\ell}X_{\ell}
-
\bigl(M_{\ell}\odot\widehat{W}_{\ell}\bigr)X_{\ell}
\bigr\|_{2}^{2},
\label{eq:local_recon}
\end{equation}
where $W_\ell$ is the weight of layer $\ell$, $X_\ell$ is the layer input, $M_\ell$ is a binary mask, and $\widehat{W}_\ell$ denotes (optionally) updated weights. Practical LLM variants include SparseGPT (Hessian-approximated sparse regression)~\cite{frantar2023sparsegptmassivelanguagemodels}, Wanda (weight–activation products)~\cite{sun2024simpleeffectivepruningapproach}, and non-uniform schemes such as FLAP~\cite{an2023fluctuationbasedadaptivestructuredpruning} and OWL~\cite{yin2025outlierweighedlayerwisesparsity} that modulate per-layer sparsity via activation variability or outlier statistics.

\emph{(ii) Gradient-based global (first-order saliency).}
Global methods score structures with respect to a model-level loss and rank them across layers. The learning objective can be formulated as:
\begin{equation}
\small
\min_{M,\ \widehat{W}} \,
\Delta\mathcal{L}\Bigl(f\bigl(X;\,M \odot \widehat{W}\bigr), \; f(X;W)\Bigr),
\label{eqt:global_pruning}
\end{equation}
where \(f\) is the forward function, \(X\) denotes the inputs, \(W\) is the original (pre-trained) weight, \(M\) is the binary mask indicating which weights remain. Given a dataset $D$, a simple first-order Taylor saliency derived from such an objective for a parameter $w$ is $|\,\tfrac{\partial \mathcal{L}}{\partial w}\cdot w\,|$; aggregating over a structure $s$ (e.g., a head or an MLP channel) gives structural importance as
\begin{equation}
\small
I(s)\;=\;\mathbb{E}_{(x,y)\in D}\!\!\Bigl[\;
\text{Aggregation}\!\left(\left\{\bigl|\tfrac{\partial \mathcal{L}(x,y;\theta)}{\partial w}\cdot w\bigr|\right\}_{w\in s}\right)
\Bigr],
\label{eq:global_firstorder}
\end{equation}

Global pruning has been extensively studied in smaller networks such as CNNs~\cite{molchanov2016pruning}, Vision Transformers~\cite{yang2023globalvisiontransformerpruning}, and compact language models~\cite{diao2023pruningdeepneuralnetworks}, consistently outperforming local approaches~\cite{blalock2020state,diao2023pruningdeepneuralnetworks}. In LLMs, methods such as LLM-Pruner~\cite{ma2023llm}, GBLM~\cite{das2024sizegradientsshapepruning}, and GISP~\cite{wang2025localglobalrevisitingstructured} instantiate this idea with structure-level aggregation and block-wise normalization to mitigate layer-scale effects.

Although these approaches are effective on generic LLM benchmarks, recent studies~\cite{zhang2025reasoningmeetscompressionunderstanding,wan2024efficientlargelanguagemodels,sui2025stopoverthinkingsurveyefficient} show that directly applying off-the-shelf structured pruning heuristics to RLMs often leads to severe degradation on multi-step reasoning tasks. 

\noindent\textbf{Calibration data for post-training pruning.}
Post-training pruning relies on a small calibration set from which activation statistics or gradient signals are extracted to calculate the importance score for pruning. 
Standard practice adopts \emph{task-agnostic} corpora such as C4~\cite{raffel2023exploringlimitstransferlearning} or WikiText~\cite{merity2016pointer} and uses prefill-style reconstruction losses to approximate layer outputs or model likelihoods~\cite{frantar2023sparsegptmassivelanguagemodels,sun2024simpleeffectivepruningapproach,ma2023llm,an2023fluctuationbasedadaptivestructuredpruning}. 
However, recent analyses~\cite{wang2025localglobalrevisitingstructured,bandari2024c4datasetoptimalpruning} on LLMs reveal that such generic calibration can severely misestimate importance and lead to sizable performance degradation. 
These observations suggest that the choice of calibration data is a key yet underexplored factor in current post-training pruning, especially for RLMs, which are more sensitive to pruning.

This work addresses that gap: we develop a post-training structured pruning framework for RLMs that explicitly focuses on calibration data design and alignment, ensuring that importance estimates reflect decode-time reasoning rather than generic language-modeling signals.

\begin{figure*}[t]
\centering
\resizebox{1\linewidth}{!}{
\includegraphics{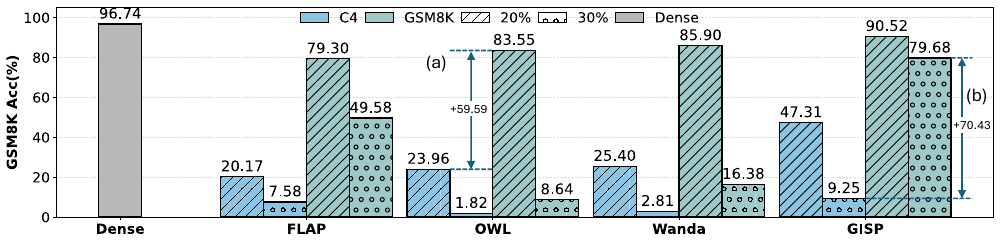}
}
% \vspace{-2.5em}
\caption{Calibration source matters for pruning reasoning LLMs. \textbf{(a)} \textbf{Observation 1:} calibrating on a \textit{generic corpus (C4)} causes severe degradation, while \textit{task-specific (GSM8K)} calibration preserves accuracy. \textbf{(b)} \textbf{Observation 2:} Gradient-based method (GISP) benefits more from task-specific calibration compared with OBS-based methods (Others).}
\label{fig:observation_exp}
\end{figure*}

\section{Observations}\label{sec:motivation}

To enable effective structured pruning for RLMs, we begin by conducting a comprehensive empirical study examining how different calibration data and pruning methods affect RLMs' reasoning performance. In practice, we examine four OBS-based pruning methods (FLAP, OWL, Wanda, SparseGPT) and one gradient-based method (GISP) on the reasoning-centric RLM Qwen3-8B. By measuring their performance on the GSM8K benchmark, we obtain three key observations:

\noindent\textbf{Observation 1} \textit{Generic calibration (e.g., C4) severely degrades reasoning abilities while task-specific calibration (e.g., GSM8K) preserves accuracy at low sparsity (e.g., 20\%)}

In post-training pruning for non-reasoning LLMs, a common setup is to use generic text corpora such as C4~\cite{raffel2023exploringlimitstransferlearning} or WikiText~\cite{merity2016pointer} for calibration and evaluate zero-shot accuracy on standard commonsense benchmarks. However, as shown in Fig.~\ref{fig:observation_exp}(a), calibrating pruning with C4 consistently leads to large accuracy drops on GSM8K across all different pruning methods. For example, at 30\%, the accuracy degraded to less than 10\%. A likely cause is the context gap between the generic calibration data and the evaluation data. To examine this hypothesis, we calibrate using a small training subset of GSM8K and observe that accuracy at low sparsity (e.g., 20\%) improves dramatically and consistently across all pruning methods. This indicates that task-specific calibration is crucial for preserving reasoning performance.

\noindent\textbf{Observation 2} \textit{Gradient-based
pruning methods benefit substantially from task-specific calibration compared to OBS-based methods}

While task-specific calibration helps preserve accuracy at low sparsity levels (e.g., 20\%), we find that its benefit diminishes for OBS-based pruning methods (e.g., Wanda, FLAP, and OWL) as sparsity increases. As shown in Fig.~\ref{fig:observation_exp}(b), these methods exhibit only marginal accuracy improvements at moderate sparsity (e.g., 30\%) and still remain far below the dense baseline. In contrast, the gradient-based method GISP continues to recover a significant fraction of reasoning accuracy under the same sparsity. This divergence stems from the different optimization objectives underlying the two classes of methods, as discussed in Sec.~\ref{sec:related_works}. Gradient-based methods explicitly rely on gradients of the model’s final loss, whose importance scores shift substantially when the calibration data better match the target reasoning task. OBS-based methods, however, focus on layer-wise activation reconstruction, which is comparatively less sensitive to the choice of calibration data. Consequently, task-specific calibration yields much larger gains for gradient-based pruning than for OBS-based pruning.

\noindent\textbf{Observation 3} \textit{Calibration with self-generated reasoning output sequences outperform task-specific calibration}

Given the decode-time nature of reasoning, we build calibration data from the model’s own self-generated solution traces using GSM8K question prompts. As shown in Fig.~\ref{fig:self-generated_and_drop_after_high_sparsity}(a), this self-generated calibration substantially improves the performance of the gradient-based GISP pruning method especially at 40\% sparsity ratio, yet yields almost no benefit for OBS-based methods. Note that other OBS-based methods, such as FLAP and OWL, exhibit similar performance to Wanda. This difference stems from how each method uses calibration data: gradient-based pruning relies on calibration distributions to compute decode-time gradients. Hence, aligning calibration with the model’s actual inference behavior produces more accurate importance scores. In contrast, OBS-based methods estimate sensitivity primarily through local Hessian approximations or activation statistics that are largely insensitive to the choice of calibration samples, making them less responsive to distribution changes. Moreover, the advantage of a self-generated output trace is further rooted in the training dynamics of reasoning LLMs. Unlike supervised fine-tuning models that learn from fixed human-written labels, RLMs are optimized through reinforcement learning on their own generated sequences, resulting in a reasoning distribution that is inherently self-induced. Consequently, self-generated calibration traces better match the model’s inference-time distribution, enhancing gradient-based pruning but providing little benefit to OBS-based methods.

\section{Self-reflective structured pruning for RLMs}\label{sec:pruning_method}

Guided by the observations presented in the previous section, we introduce RESP, a self-reflective structured pruning framework tailored for reasoning LLMs. RESP consists of three key components: (1) \textbf{self-reflective calibration}, which constructs self-generated, task-specific reasoning traces that better match the model’s inference-time distribution; (2) \textbf{gradient-based importance estimation}, which computes decode-only, gradient-based importance scores to guide structured pruning; and (3) \textbf{progressive regeneration calibration}, which refreshes calibration traces during pruning to maintain alignment and preserve performance as sparsity increases.

% Guided by this principle, we instantiate a self-reflective pruning pipeline for reasoning LLMs. It has two components: (i) a \emph{calibration and importance} module that builds a task-specific pool of \emph{self-generated} decode-time traces and computes decode-only gradient saliency; and (ii) \emph{progressive regeneration} that refreshes calibration traces at pruning milestones to counter calibration drift observed at high sparsity.

\subsection{Self-reflective calibration}\label{sec:calib_imp}

% \subsection{Calibration and importance estimator}\label{sec:calib_imp}

Motivated by \textbf{Observation 3}, the calibration signal should be derived from the model’s \emph{own} decode-time reasoning trace rather than generic corpora or human-written rationales. Specifically, given a task-specific prompt question $x$, a max context length $T$, and a max calibration sequence length $L$, the RLM produces a chain-of-thought and final answer $(y_{0:T-1})$ under its standard decoding procedure. We extract the decode output segment $\tau=(y_{0:L-1})$ and construct a calibration set
$\hat{\mathcal{D}}=\{(x,\tau)\}$ as \emph{self-generated} traces.

\subsection{Gradient-based importance estimation}
Our \textbf{Observations 2 and 3} show that gradient-based pruning methods benefit substantially from task-specific calibration, making them particularly effective for RLMs when paired with self-generated calibration. To leverage this, RESP estimates structural importance using decode-only, gradient-based importance score, ensuring that pruning decisions are guided by signals aligned with the model’s actual reasoning process.

\noindent\textbf{Decode-only loss for reasoning alignment.}
In practice, given a calibration pair $(x, \tau)$ (e.g, a GSM8K problem), we mask out prefill tokens and evaluate loss only on tokens produced during the decode phase. Let $m_t\in\{0,1\}$ indicate whether token $t$ belongs to the decode phase and $Z=\sum_t m_t$. The decode-time negative log-likelihood is
\begin{equation}
\small
\mathcal{L}_{\mathrm{dec}}(x,\tau;\theta)
= \frac{1}{Z}\sum_{t} m_t \Big(-\log p_\theta(y_t \mid x, y_{<t})\Big).
\end{equation}
Unlike full-sequence or prefill-dominated objectives, this loss isolates the token-by-token reasoning steps that directly determine task performance, preventing prefill reconstruction terms from distorting importance estimation.

\noindent\textbf{Gradient-based importance score for structured pruning.} We then exploit the data sensitivity of gradient-based pruning by computing a first-order, decode-only importance on model-generated traces. For a parameter $w$, the importance score based on first-order Taylor saliency is defined as
\begin{equation}
\small
I(w)=\mathbb{E}_{(x,\tau)\sim\hat{\mathcal{D}}}
\left[\left|\frac{\partial \mathcal{L}_{\mathrm{dec}}(x,\tau;\theta)}{\partial w}\odot w\right|\right].
\end{equation} Afterward, following standard gradient-based pruning practice \cite{wang2025localglobalrevisitingstructured,yang2023globalvisiontransformerpruning}, we first aggregate weight-level importance by summing it over predefined structures (e.g., attention heads or MLP channels), normalize the aggregated structural importance, rank all structures across the entire model, and finally prune the least important structures to meet the target sparsity.

% \paragraph{Decode-only masking}

% To align importance estimation with inference, we mask out prefill tokens and define a \emph{decode-only} sample loss. Let $m_t\in\{0,1\}$ indicate whether token $t$ belongs to the decode phase and $Z=\sum_t m_t$. The decode-time negative log-likelihood is
% \[
% \mathcal{L}_{\mathrm{dec}}(x,\tau;\theta)
% = \frac{1}{Z}\sum_{t} m_t \Big(-\log p_\theta(y_t \mid x, y_{<t})\Big).
% \]
% This removes prefill-dominated reconstruction effects and centers the calibration signal on the token-by-token decisions that govern reasoning.

% \paragraph{Reasoning-centric gradient importance}
% We then exploit the data sensitivity of gradient-based pruning by computing a first-order, decode-only importance on model-generated trajectories. For a parameter $w$, the first-order Taylor saliency is
% \[
% I(w)=\mathbb{E}_{(x,\tau)\sim\hat{\mathcal{D}}}
% \left[\left|\frac{\partial \mathcal{L}_{\mathrm{dec}}(x,\tau;\theta)}{\partial w}\odot w\right|\right],
% \]
% then we perform structured aggregation, ranking, and iterative pruning under the target sparsity schedule following common pruning practice.

\subsection{Progressive calibration regeneration}

\begin{figure}[tbp]
\centering
\resizebox{1\linewidth}{!}{
\includegraphics{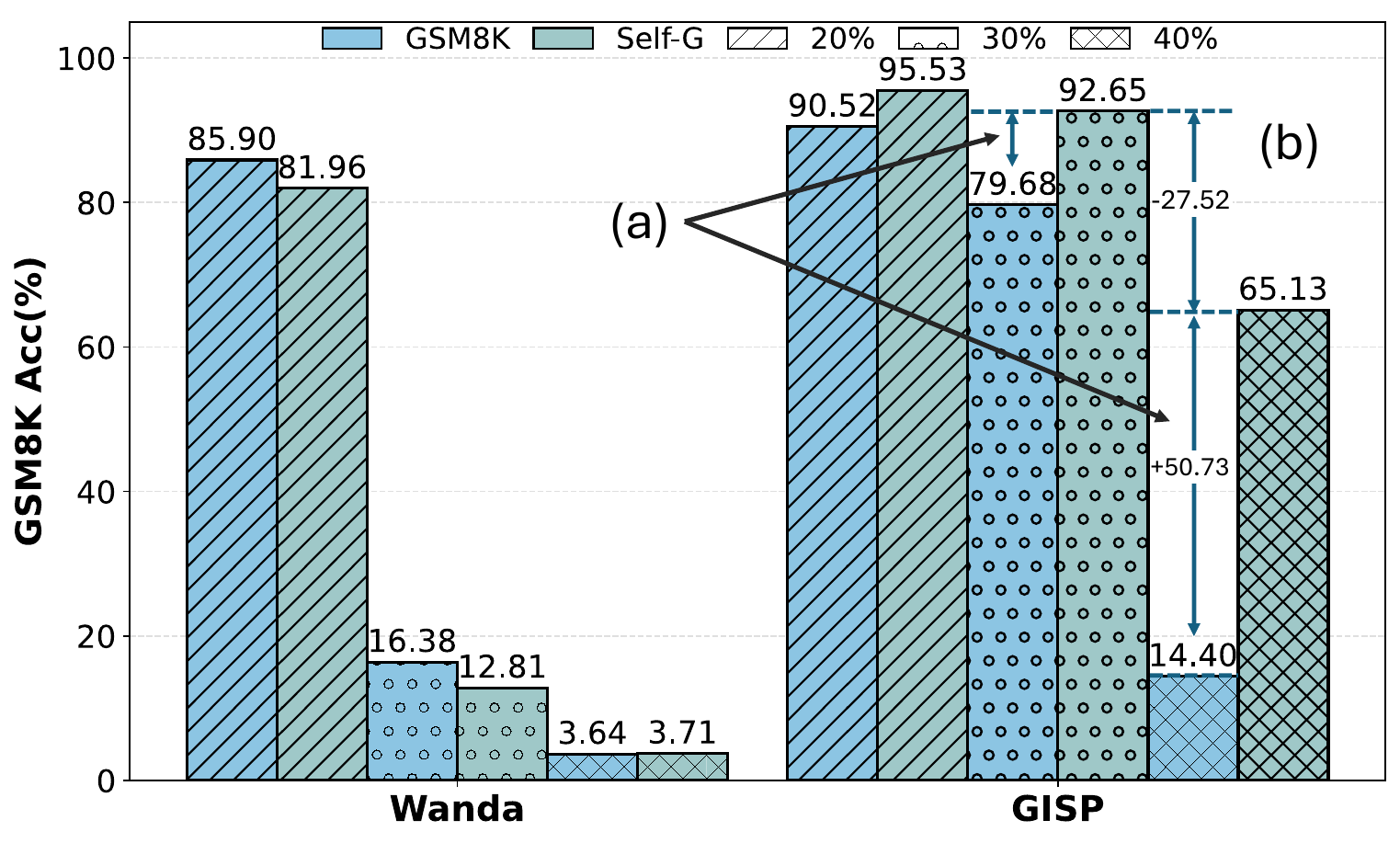}
}
\caption{Illusration of \textbf{(a) Observation 3:} Calibration with self-generated reasoning traces outperform task-specific calibration. Furthermore, (b) the pruned model still has a significant performance drop after entering the high sparsity domain, motivating us to develop progressive regeneration. "Self-G" marks for self-generated calibration data.}
\label{fig:self-generated_and_drop_after_high_sparsity}
\end{figure}

Fig.~\ref{fig:self-generated_and_drop_after_high_sparsity}(b) reveals a pronounced accuracy cliff around 40\% sparsity when pruning is driven solely by dense-model calibration traces. Although such traces initially align well with the model’s reasoning behavior, increasing sparsity progressively alters decode-time traces, making earlier traces distributionally stale. As a result, importance estimates increasingly reflect outdated gradients, leading to misinformed pruning decisions. This calibration drift highlights the need to realign calibration with the model’s evolving reasoning distribution as sparsity grows.

% We further observe that self-generated calibration collected from the dense model becomes mismatched at high sparsity (e.g., 40\%) due to a shift in the model’s reasoning distribution. To address this, RESP periodically regenerates fresh self-generated reasoning traces from the current pruned model as sparsity increases, keeping calibration aligned with the evolving model and preventing drift that would otherwise accumulate across pruning iterations (Fig.\ref{fig:overview}(c)).

% As shown in Fig.~\ref{fig:self-generated_and_drop_after_high_sparsity}, when entering the high sparsity domain, there is still a significant performance drop. 

% We attribute this to \emph{calibration drift}: increasing structured sparsity perturbs decode-time reasoning traces, so those collected at the dense model become mismatched with the current model, leading to failure in importance estimation.

\noindent\textbf{Progressive calibration regeneration at pruning milestones.}
To address the calibration drift, we further propose an improved technique \textbf{progressive regeneration}: at pre-defined pruning milestones 
$\mathcal{A}=\{\alpha_1,\alpha_2,\dots,\alpha_K\}$ with $\alpha_K$ the target sparsity, we entirely replace the previously collected calibration traces with newly generated ones by the \emph{current} model $\theta^{(\alpha_s)}$ on the fixed prompt set $\mathcal{X}$ at current milestone $\alpha_{s}$. 
We then compute the decode-only, gradient-based importance on the refreshed set 
$\hat{\mathcal{D}}^{(\alpha_s)}=\{(x,\tau^{(\alpha_s)})\,|\,x\in\mathcal{X}\}$ and prune to the next milestone $\alpha_{s+1}$. It is worth noting that during regeneration, the token budget and prompt set are kept fixed across sparsity milestones. Only the decode-time traces are refreshed using the current pruned model, ensuring that updates arise solely from aligning calibration to the model’s evolving behavior.
% Any improvement arises solely from \emph{distributional alignment} (sampling from $P_{\theta^{(s)}}$): both the number of prompts $|\mathcal{X}|$ and the per-trace token cap are held fixed across stages, and no external new information (e.g., correctness of answer) is provided.
Empirically, this alignment yields smoother degradation curves and mitigates large drops around the critical milestone ratio at high sparsity.

\section{Experiments}
\subsection{Experiments Setup} 

\noindent\textbf{Model and Tasks.}
We primarily evaluate on \textbf{Qwen3-8B}~\cite{yang2025qwen3technicalreport} as the target model. 
For reasoning benchmarks, we choose two representative math word–problem datasets:
\textbf{MathQA} (multiple-choice math problems and we report Top-1 choice accuracy)~\cite{amini2019mathqainterpretablemathword} and \textbf{GSM8K} (free-form grade-school math and we report Top-1 exact-match on the final answer)~\cite{cobbe2021trainingverifierssolvemath}.
Following common practice, MathQA is evaluated in \emph{zero-shot} with a short instruction prompt, while GSM8K uses an \emph{8-shot} prompt. The benchmarks are applied through the EleutherAI LM Harness pipeline~\cite{eval-harness}. Both \emph{evaluation} and \emph{calibration trace generation} use the provider-recommended sampling setup:
\texttt{do\_sample=True}, \texttt{temperature=0.6}. Evaluation \emph{max\_new\_tokens} is 4096 for MathQA and 2048 for GSM8K. For robustness to formatting, we use Qwen3-30B-A3B-Instruct-2507 as a light-weight answer extractor to select the final option (MathQA) or extract a final scalar (GSM8K), as regex normalization might be ambiguous for models without the output alignment fine-tuning process. 
Empirically, the extractor contributes at most $\pm$0.7 percentage points variation; we hold it fixed across all methods for fairness. We fix the random seed to 0 for prompt sampling and decoding parameters across methods. All calibration datasets share the same $\mathcal{X}$, consisting of random $500$ prompts from the training split per task. When generating calibration traces for RESP, we cap the decode length at $2048$ for all tasks.

% For all self‑generated calibration runs, importance is computed with decode‑only masking. Wanda does not provide such option for masking.

% \paragraph{Calibration pools}
% To probe the role of calibration, we consider:

% \vspace{-1em}
\noindent\textbf{Baselines.} We compare against two representative structured pruning baselines: \textbf{Wanda}~\cite{sun2024simpleeffectivepruningapproach} and \textbf{GISP}~\cite{wang2025localglobalrevisitingstructured}.
\textbf{Wanda} is an OBS-based method that ranks structures by the magnitude of weight–activation products, leveraging the outlier phenomenon raised in LLMs.
\textbf{GISP} is a post-training, gradient-based global iterative approach that removes attention heads and MLP channels using first-order gradients aggregated at the structure level with block-wise normalization.
For a fair comparison, we adopt the same stability constraint across all methods: we do not prune the first 10\% of Transformer blocks, and we keep the final block unpruned.

% \paragraph{Implementation and environment}
% We implement our pipeline on top of the public \textsc{GISP} codebase, adding (i) self-reflective calibration with decode-only masking and (ii) progressive regeneration logic. The regenerate ratio $\rho$ is set to be 1. We fix the random seed to 0 for prompt sampling and decoding parameters across methods. For all self‑generated calibration runs (both GISP and Ours), importance is computed with decode‑only masking (prefill tokens are masked). Wanda, as an activation‑based method, is forward‑only and has no decode‑only counterpart. Unless otherwise noted, all experiments are conducted on a cloud computing server with an AMD EPYC 9554 CPU, 512 GB of memory, 400GB SSD, and 8 Nvidia H200 141GB GPUs. 

\subsection{Experiments Results.}

\begin{table}[t]
\centering
% \vspace{-1em}
\caption{Main results on Qwen3-8B.}
% \vspace{-1em}
\label{tab:qwen3_main}
\resizebox{0.85\linewidth}{!}{
\begin{tabular}{c|l|c|c}
\toprule
\textbf{Pruning Ratio} & \textbf{Method}  & \textbf{GSM8K} & \textbf{MathQA} \\
\midrule
0\%  & Dense & 96.74\% & 87.60\% \\
\midrule
\multirow{5}{*}{20\%} 
 & Wanda           & 85.90\% & 51.42\% \\
 % & Wanda & Self-generated  & 81.96\% & 81.11\% \\
 & GISP       & 90.52\% & 73.00\% \\
 & RESP w/o re-gen  & \textbf{95.53\%} & \textbf{84.02\%} \\
 & RESP  & \textbf{95.53\%} & 83.69\% \\
\midrule
\multirow{5}{*}{30\%} 
 & Wanda            & 16.38\% & 11.22\% \\
 % & Wanda & Self-generated  & 12.81\% & 30.89\% \\
 & GISP             & 79.68\% & 27.07\% \\
 & RESP w/o re-gen  & \textbf{92.65\%} & 80.30\% \\
 & RESP   & 91.36\% & \textbf{81.54\%} \\
\midrule
\multirow{5}{*}{40\%} 
 & Wanda            & 3.64\%  & 5.73\% \\
 % & Wanda & Self-generated  & 3.71\%  & 9.21\% \\
 & GISP             & 14.40\% & 12.60\% \\
 & RESP w/o re-gen    & 65.13\% & 16.78\% \\
 & RESP   & \textbf{81.27\%} & \textbf{59.60\%} \\
% \midrule
% \multirow{5}{*}{50\%} 
%  & Wanda & Gold            & 0.00\%  & 5.83\% \\
%  & Wanda & Self-generated  & 0.00\%  & 2.14\% \\
%  & GISP  & Gold            & 0.00\%  & 6.33\% \\
%  & GISP  & Self-generated  & 6.82\%  & 4.72\% \\
%  & Ours  & Self-generated  & \textbf{17.29\%} & \textbf{9.55\%} \\
\bottomrule
\end{tabular}}
% \vspace{-2em}
\end{table}

\noindent\textbf{The effectiveness of self-generated calibration data.} Presented in Table~\ref{tab:qwen3_main}, compared with other baselines, \textbf{RESP w/o re-gen} substantially improves gradient-based pruning across pruning ratios. For GISP, 5–13\% on GSM8K and 11–53 \% on MathQA gain can be observed at 20–30\% sparsity. At 40\% sparsity, the trend persists on GSM8K, with 51\% improvement, while gains on MathQA taper off, suggesting the need for regeneration.

% In contrast, Wanda shows mixed, task‑dependent sensitivity to the calibration source. These results support our hypothesis that \emph{the calibration plays a larger and more consistent role for gradient‑based importance}: gradient signals align better when derived from the model’s own decode‑time trajectories, whereas OBS‑based heuristics exhibit smaller and less stable gains across tasks and sparsity levels

\noindent\textbf{The effectiveness of regeneration.} Furthermore, comparing between \textbf{RESP w/o re-gen} to \textbf{RESP} isolates the performance gain of regeneration. As shown in Table~\ref{tab:qwen3_main}, at low sparsity, regeneration yields modest gains: both variants obtain similar performance. We hypothesize that, in this regime, the benefit from better distribution alignment is partly offset by the degradation in generation quality induced by pruning: the regenerated data are better matched to the sparse model’s behavior, but are also of slightly lower quality than those produced by the dense model, leading to a limited net improvement. When sparsity is high, regeneration markedly delays collapse: on GSM8K, we achieve an improvement of 16.14\% at 40\%; on MathQA, the gains are even larger, up to 42.82\% at 40\%. This suggests that, in the high-sparsity regime, the distribution gap becomes the dominant factor, and aligning the calibration data with the sparse model via progressive regeneration is crucial for maintaining performance. Overall, progressive regeneration yields smoother degradation curves and preserves usable accuracy well beyond the critical sparsity regime.

\begin{table}[t]
\centering
% \vspace{-1em}
\caption{Effect of regeneration on MathQA accuracy under different pruning milestones. }
% \vspace{-1em}
\label{tab:ab_regen_path}
\resizebox{1\linewidth}{!}{
\begin{tabular}{c|c|c|c|c}
\toprule
\textbf{\parbox{1.3cm}{\textbf{Milestone\\Ratio}}} & \textbf{\parbox{1.9cm}{\textbf{Next Target\\Ratio}}}  & \textbf{Regenerate?} & \textbf{MathQA (Calibration)} & \textbf{MathQA (Test)} \\
\midrule
\multirow{2}{*}{10\%} & \multirow{2}{*}{20\%} & F & 90.60\% & 86.47\% \\
 &  & T & 90.00\% & 86.47\% \\
\midrule
\multirow{2}{*}{20\%} & \multirow{2}{*}{30\%} & F & 84.80\% & 81.54\% \\
 &  & T & \textbf{88.00\%} & \textbf{84.92\%} \\
\midrule
\multirow{2}{*}{30\%} & \multirow{2}{*}{40\%} & F & 38.40\% & 40.60\% \\
 &  & T & \textbf{58.00\%} & \textbf{56.35\%} \\
\midrule
\multirow{2}{*}{40\%} & \multirow{2}{*}{50\%} & F & 5.60\% & 7.24\% \\
 &  & T & \textbf{9.80\%} & \textbf{10.08\%} \\
\bottomrule
\end{tabular}}

\end{table}

\subsection{Ablation}
\noindent\textbf{Necessity of regeneration.} To further analyze the necessity and effect of regeneration at each pruning milestone, we perform path-contrast experiments starting from the same checkpoint, pruning to the next sparsity with and without refreshing the calibration traces.
As shown in Table~\ref{tab:ab_regen_path}, regeneration (\textbf{T}) consistently leads to higher accuracy on both test set and calibration sample set across all milestone intervals compared with no regeneration (\textbf{F}), confirming its necessity for maintaining reasoning capability as sparsity increases.
Notably, the performance gap widens in higher sparsity regimes (e.g., $+15.75\%$\, from 30\%$\to$40\%), where calibration drift becomes more pronounced.
These results justify adopting full regeneration at every milestone as the default configuration.

\begin{table}[h]
% \vspace{-1em}
\centering
\caption{Comparison of different regeneration ratios}
% \vspace{-1em}
\label{tab:ab_regenerate_ratio}
\resizebox{0.9\linewidth}{!}{
\begin{tabular}{c|c|c|c|c}
\toprule
\textbf{\parbox{1.3cm}{\textbf{Milestone\\Ratio}}} & 
\textbf{\parbox{1.9cm}{\textbf{Next Target\\Ratio}}} & 
\textbf{\parbox{1.5cm}{\textbf{Regenerate\\Ratio}}} & 
\textbf{GSM8K} & 
\textbf{Improvement} \\
\midrule
\multirow{6}{*}{30\%} & \multirow{6}{*}{40\%} & 0 & 65.13\% & 0\% \\
 &  & 0.2 & 71.11\% & +5.99\% \\
 &  & 0.4 & 75.28\% & +10.16\% \\
 &  & 0.6 & 75.89\% & +10.77\% \\
 &  & 0.8 & 82.56\% & +17.44\% \\
 &  & 1 & 83.17\% & +18.04\% \\
\bottomrule
\end{tabular}
}
\end{table}

\noindent\textbf{Different regeneration ratio.} To analyze the effect of regeneration ratios, we pause the iterative pruning process at 30\%, vary the regeneration ratio \(\rho\in[0,1]\), replacing a \(\rho\)-portion of the calibration traces with those regenerated, while keeping the calibration budget and prompts fixed. Then we utilize different regenerated calibration mixes to continue pruning to 40\%. The results are presented in Table~\ref{tab:ab_regenerate_ratio}, which showing that the accuracy on GSM8K increases \emph{monotonically} with \(\rho\): from 65.13\% at \(\rho{=}0\) (no refresh) to 83.17\% at \(\rho{=}1\) (fully regenerate, \(+18.04\%\)). Thus, we adopt full regeneration as the default.

\section{Conclusion}
In this work, we show that conventional post-training pruning pipelines fail on reasoning LLMs due to a mismatch between calibration data, pruning objectives, and decode-time reasoning behavior. By introducing RESP, a self-reflective structured pruning framework built on self-generated calibration traces, decode-only gradient-based importance, and progressive regeneration, we align pruning decisions with the model’s actual inference distribution. Experiments on Qwen3-8B demonstrate that RESP preserves near-dense accuracy at 20–30\% sparsity and significantly mitigates collapse at higher sparsity, establishing a practical and effective path toward compressing reasoning-centric language models.

% We revisited post-training structured pruning for reasoning LLMs and traced its brittleness to two compounding mismatches, calibration and objective. We introduced RESP, a post‑training framework that calibrates with the model’s self‑generated decode‑time traces, computes decode‑only gradient saliency for structural components, and progressively regenerates traces at pruning milestones to counter calibration drift. On Qwen3‑8B across GSM8K and MathQA, RESP remains near‑dense at moderate sparsity and substantially delays the high‑sparsity cliff, outperforming baselines under the same budget. The overarching lesson is simple: calibrate and prune in the distribution you will decode in via a decode‑aligned, self‑generated calibration, plus regeneration, is a practical recipe for compressing reasoning models while preserving their utility.
\newpage

%\balance
\bibliographystyle{ACM-Reference-Format}
\bibliography{mybibfile}

\end{document}